\documentclass[
twocolumn,
]{ceurart}

\usepackage{xurl}
\usepackage{listings}
\usepackage{todonotes}
\begin{document}

\copyrightyear{2023}
\copyrightclause{Copyright for this paper by its authors.
  Use permitted under Creative Commons License Attribution 4.0
  International (CC BY 4.0).}

\conference{preprint, November 2022}

\title{A taxonomic system for failure cause analysis of open source AI incidents}


\author[1]{Nikiforos Pittaras}[%
email=pittarasnikif@gmail.com,
]
\cormark[1]
\address[1]{Responsible AI Collaborative}

\author[1]{Sean McGregor}[%
email=smcgregor@seanbmcgregor.com,
]


\cortext[1]{Corresponding author.}

\begin{abstract}
While certain industrial sectors (e.g., aviation) have a long history of mandatory incident reporting complete with analytical findings, the practice of artificial intelligence (AI) safety benefits from no such mandate and thus analyses must be performed on publicly known ``open source'' AI incidents. Although the exact causes of AI incidents are seldom known by outsiders, this work demonstrates how to apply expert knowledge on the population of incidents in the AI Incident Database (AIID) to infer the potential and likely technical causative factors that contribute to reported failures and harms.
We present early work on a taxonomic system that covers a cascade of interrelated incident factors, from system goals (nearly always known) to methods / technologies (knowable in many cases) and technical failure causes (subject to expert analysis) of the implicated systems. We pair this ontology structure with a comprehensive classification workflow that leverages expert knowledge and community feedback, resulting in taxonomic annotations grounded by incident data and human expertise.
\end{abstract}


\maketitle

\section{Introduction}


The pursuit of safe AI is a critical problem of the 21st century which, unless dealt with, harbors dangers likely to define and shape the trajectory and prosperity of the human species \cite{yudkowski2022agi}.
AI Safety research efforts have made progress towards combating AI-induced x-risk in a variety of fronts; these include value loading and refinement by human preferences \cite{ouyang2022training}, investigations on inner misalignment manifestations \cite{cobbe2019quantifying}, interpretability-oriented analysis on deep network models \cite{elhage2021mathematical}, as well as conceptual work on frameworks and potential harms of superintelligent systems \cite{drexler2019reframing,yampolskiy2021agi}.
In contrast, there has been limited work in a different direction: exploiting publicly available data
that may provide useful insights in the function, composition, alignment, deployment and application failure causes of real-world AI systems.
A prominent example of such data streams are AI incidents, i.e. public articles and reports that describe harms and failures of deployed AI systems in the wild.
This study presents early work, proposing the
analysis and annotation of AI incidents via the development of a taxonomic system that captures Goals, Methods / Technologies and Failure Causes of a technical nature (abbreviated as GMF), that stem from content-based information in incident reports in conjunction with technical knowledge and expertise in the AI Safety and Alignment community.
In the rest of this paper we provide the motivation and contributions of preliminary work conducted to develop the taxonomic system, including the proposed structure, annotation workflow and development procedures. Finally, we discuss the expected impact of GMF towards the annotation, discovery and analysis of AI systems and their failure causes as they manifest in existing incidents, along with the potential of resulting datasets for future data-driven AI Safety and Alignment research efforts.


\section{Related Work} \label{sec:rw}

While research efforts have been focused on analyzing and categorizing failures \cite{yampolskiy2016artificial,banerjee2020ai,yampolskiy2018predicting,mallah2017landscape,manheim2018categorizing} and major components of AI systems \cite{honavar2006artificial,samoili2020ai}, there has been limited work on holistically linking multiple aspects of AI systems into a single, interrelated taxonomic model. Additionally, these efforts are often separated from real-world failures and events, resulting in the under-utilization of any research outputs and taxonomic insights via, e.g., shared datasets of practical utility. Finally, when applying existing taxonomies to real world systems, they can rarely be applied adequately with incomplete or uncertain information, which is the case for the vast majority of AI-related incidents now reported.

Research efforts in the AI safety community have produced resources to organize different aspects of the current landscape that builds towards safe AI, encompassing organizations, datasets, failures, and harms \cite{ai4good,zheng2018cybersecurity,yampolskiy2016artificial,mallah2017landscape}.
Notable works include the CSET taxonomy \cite{cset}, which provides a broad set of information for AI incident annotation, ranging from high-level descriptions of harm types, severity estimates and distribution among different affected groups, to limited sets of high-level AI functions, causative factors (e.g. robustness failure) and a variety of additional metadata (e.g. system owner, deployment sector, user expertise etc.). While rich, the ontology aims for broad descriptions of incidents and their real-world impact, rather than focusing on the technical aspects and failure causes of the AI system involved.
Other efforts focus on the compilation of a chronological lists of sourced AI Failures \cite{yampolskiy2016artificial}, paired with a comparison between AI Safety and cybersecurity viewpoints and concerns around producing safe and reliable systems. The provided incident pool is however limited ($< $ 20), with no analysis performed on the specific incidents cited.

Some research efforts offer causal factors for AI misbehavior \cite{banerjee2020ai}, paired with harm descriptions and linked to exemplary real-world manifestations, along with general directions for mitigation and handling per category. Although informative, the technical grounding of causal factors is limited, with the majority of the discussion being delivered at a conceptual, high-level framework and no additional views (e.g. categorizations of system objectives or implementation descriptions) being considered.
Further recent research focuses on specific domains and models, such as the taxonomy of language model risks \cite{weidinger2022taxonomy}, which categorizes real-world harms, risks and hazards of large language generators; while the study is comprehensive and provides technical mitigation approaches, the proposed taxonomic grouping focuses on a very high-level view of harmful effects and is restricted to the language domain, without aiming to explore causative factors that potentially generalize across systems and tasks of different modalities.

In light of these efforts, this work describes ongoing work on the proposed Goals, Methods and Failure Causes taxonomic system, which encapsulates three interrelated ontologies: 1) high-level AI system goals, 2) methods and technologies used for system implementation, and 3) technical failure causes that result in misbehavior in the applied system;
this structure utilizes inter-taxonomy relationships towards identifying technical failure factors in AI systems. The resource is presented in the context of characterizing real-world AI incidents provided by the AI Incident Database (AIID) \cite{mcgregor2021preventing}.

The list of contributions associated with this preliminary body of work includes:


\begin{itemize}
\item General, holistic, interrelated taxonomies: We propose three interconnected views of a broad and diverse set of AI incidents, enabling multi-faceted data interpretation, analysis and retrieval, yielding various pattern matching avenues towards diagnosing and mitigating harms by parties of different interests, domain and expertise.
\item Focus on technical causal factors: While most existing works strive to categorize failures and harms, we focus on AI attributes, approaches, limitations and issues of a technical nature, that themselves lead to observed harms.
\item Grounded to real data: In contrast to conceptual / theoretical analysis or narrow experimental testbeds prone to measurement issues \cite{wentworth2022measuring}, our study centers around annotating real-world AI incidents provided by AIID, which describe real-world observations of AI systems and failures.
\item Data-driven, fine-grained and explainable: we propose a workflow that links annotations to specific incident text spans and metadata, enhancing explainability, transparency and validation and enabling
  further data-driven safety research.
\end{itemize}

Given this setting, we move on to provide a description of the structure, workflows, development procedures and expected impact of GMF in the sections that follow.

\section{The GMF taxonomic system}\label{sec:gmf}

Here we describe the proposed taxonomic system structure, annotation workflow, development procedure and projected impact of GMF, as reflected by the initial body of research work and early findings.

\subsection{Taxonomies} \label{sec:taxonomies}
Three interrelated taxonomies are included in GMF, providing a well-rounded view and different discovery avenues for AI systems involved in incidents.

First, the \textit{AI System Goals} taxonomy addresses what the deployed AI system was trying to achieve; it encapsulates high-level goals, objectives and primary use cases pursued in the real world, such as ``Translation'' or ``Face Recognition''. This information enables use case-driven incident discovery and facilitates retrieval of historical AI methods and failure causes of similar systems by interested groups, such as AI developers and safety engineers.

Second, \textit{AI Methods and Technologies} contains information on how the system is built, including learning models, representation construction approaches and other methodological, engineering and implementation-related features, e.g. ``Transformer Neural Network'', ``Collaborative Filtering''. Incident filtering by elements of this taxonomy provides popularity and utilization trends of different technologies, while highlighting the distributions of harms, historical failures and technical pitfalls associated with specific implementation approaches.

Finally, the \textit{AI Failure Causes} taxonomy consists of technical reasons that lead to the emergence of real-world harms during AI deployment. It involves systemic failures of technical nature which may manifest in system design, engineering, specifications and construction procedure, such as ``Concept Drift'' or ``Distributional Bias'', which are potential causal factors to the observed undesirable AI behavior. Failure cause-based retrieval from GMF-annotated instances should reveal indicative use cases, methods and technologies where specific failures materialize, enabling data collection for pattern extraction and causal analysis, as well as provide groundwork for research into mitigation efforts.

At the present stage, taxonomy elements represent general categories (e.g. ``Clustering'' instead of ``K-Means Clustering'') in order to establish high term applicability and are composed of a short descriptive name (e.g. 1-3 words) and a concise description (e.g., up to 30 words) that communicates exact semantic content to annotators.

%
\subsection{Annotation Workflow} \label{sec:workflow}
Given that AI incident report contents may contain limited amounts of technical information for efficient annotation with GMF,
we propose an annotation workflow that additionally leverages taxonomic relationships, historical incident records and technical knowledge in the AI, ML and Safety community.

The primary source of information available to annotators for arriving at relevant GMF incident classifications is
AI incident contents, i.e. text available in reports that describe the incident. High-level AI system information (e.g., system objective / domain / use-case) required to apply an \textit{AI System Goals} classification should be readily obtainable from incident contents, with no additional investigation and limited speculation.
Second, annotators have access to previously annotated incident collections in AIID; labelled incident retrieval provides informative priors via historical incidents, which are useful in the annotation of incidents that include limited technical information, especially for the \textit{AI Methods and Technologies} and \textit{AI Failure Causes} taxonomies, as described in section \ref{sec:taxonomies}.
Finally, knowledgeable individuals in AI, Machine Learning (ML) and AI Safety as well as other valuable disciplines (e.g. machine ethics, human-computer interaction, cybersecurity, philosophy, mathematics, etc.) can draw on training, experience and analytic skills to provide diverse, critical insight on methods, technologies and failure causes that produce events and harms mentioned in the incident, when relevant technical information is not directly available but can be inferred. Notably, such insights can be extracted via crowdsourcing means, when coverage of large amounts of incident data is prioritized, using careful moderation and curation to account for labelling noise that may result from crowdsourced annotators with different backgrounds and levels of expertise.
 
\begin{figure*}
  \centering

  \includegraphics[scale=0.2,width=1.0\textwidth]{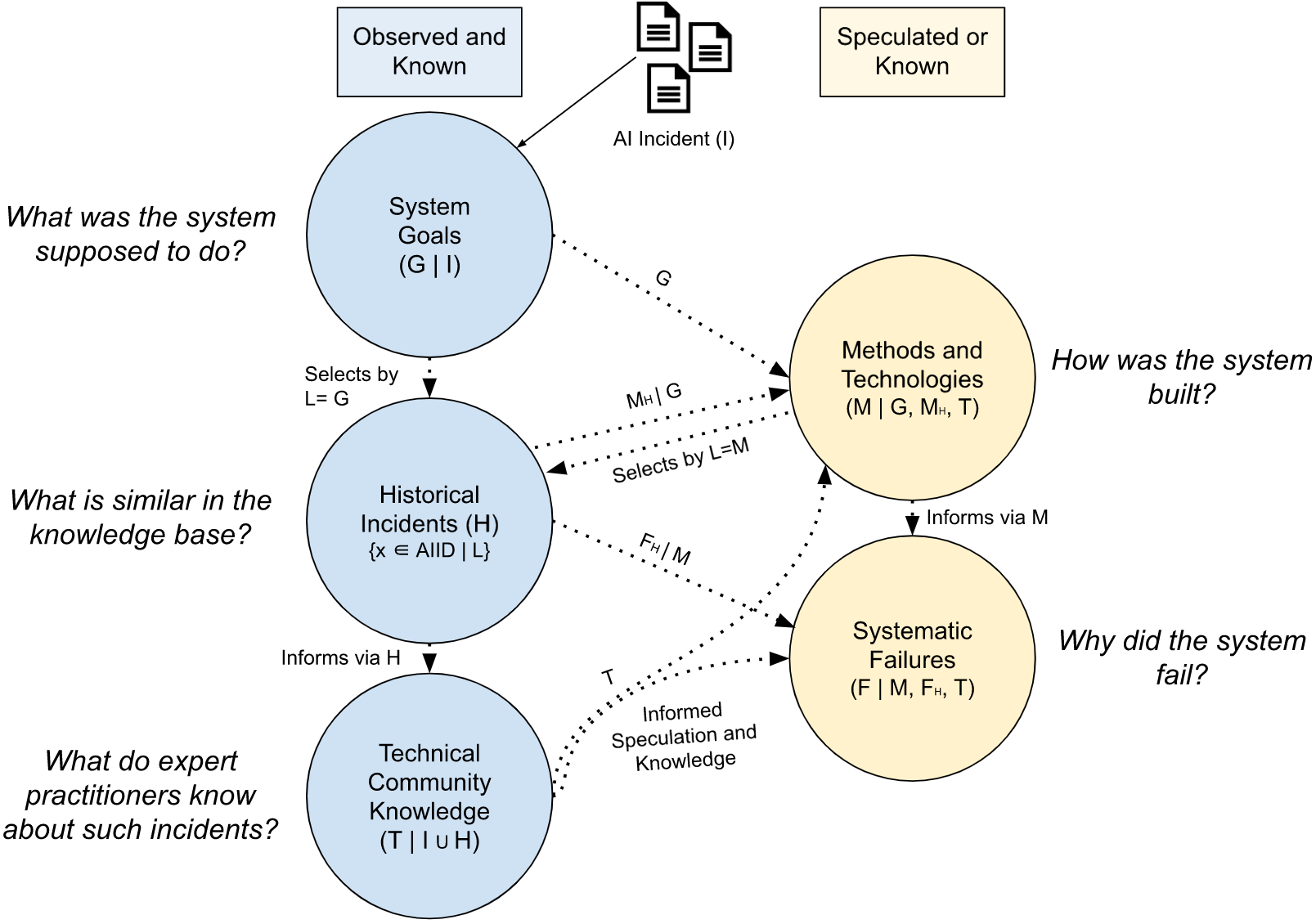}
  
  \caption{The proposed annotation workflow of AIID instances with GMF, illustrating relationships between taxonomic elements and related sources of information (AIID data, technical AI / ML / Safety community). System Goals are generally known for all incidents in the AIID. The system goal $G$ then determines which candidate similar incidents $H$ are pulled from the AIID and presented to the technical community. Then, the combination of previous methods and technologies $M_H$ from the AIID conditioned on the Goals and Technical Community Knowledge $T$ helps determine candidate Methods and Technologies (M) whenever they are not stated in available incident reports. Finally, the combination of historical incidents for similar system goals and methods help determine the candidate Systemic Failures (F) with the input of the Technical Community.}
  \label{fig:diagram}
\end{figure*}

Given these information streams, the proposed annotation workflow for an incident $I$ is the following:
\begin{enumerate}
\item Incident annotation with the \textit{AI System Goals} taxonomy; the appropriate classification $G$ should be easily identifiable by incident contents alone.
\item Retrieve similar incidents $H$ from AIID that share a goal classification $L=G$, providing a use case-based context of methods and technologies $M_H$
\item Extract relevant technical community knowledge on the incident and all available context ($I \cup H$)
\item Arrive at likely annotations $M$ from the AI Methods and Technologies taxonomy, $M | I, G, M_H, T$, i.e. by considering given incident contents, current goal classification, historical incident method classifications and relevant technical community knowledge.
\item Update the historical incident pool $H$ by considering a lookup parameter of the method / technology annotation, $L=M$.
\item Arrive at likely annotations $F$ from the AI Failure Causes taxonomy, i.e. with respect to current incident contents and classification ($I, G, M$), as well as historical failure annotations and expert knowledge from the community ($F_H, T$).

\end{enumerate}
An illustration of the proposed annotation workflow is presented in Figure \ref{fig:diagram}.

\begin{figure*}
  \centering
 \includegraphics[width=\textwidth]{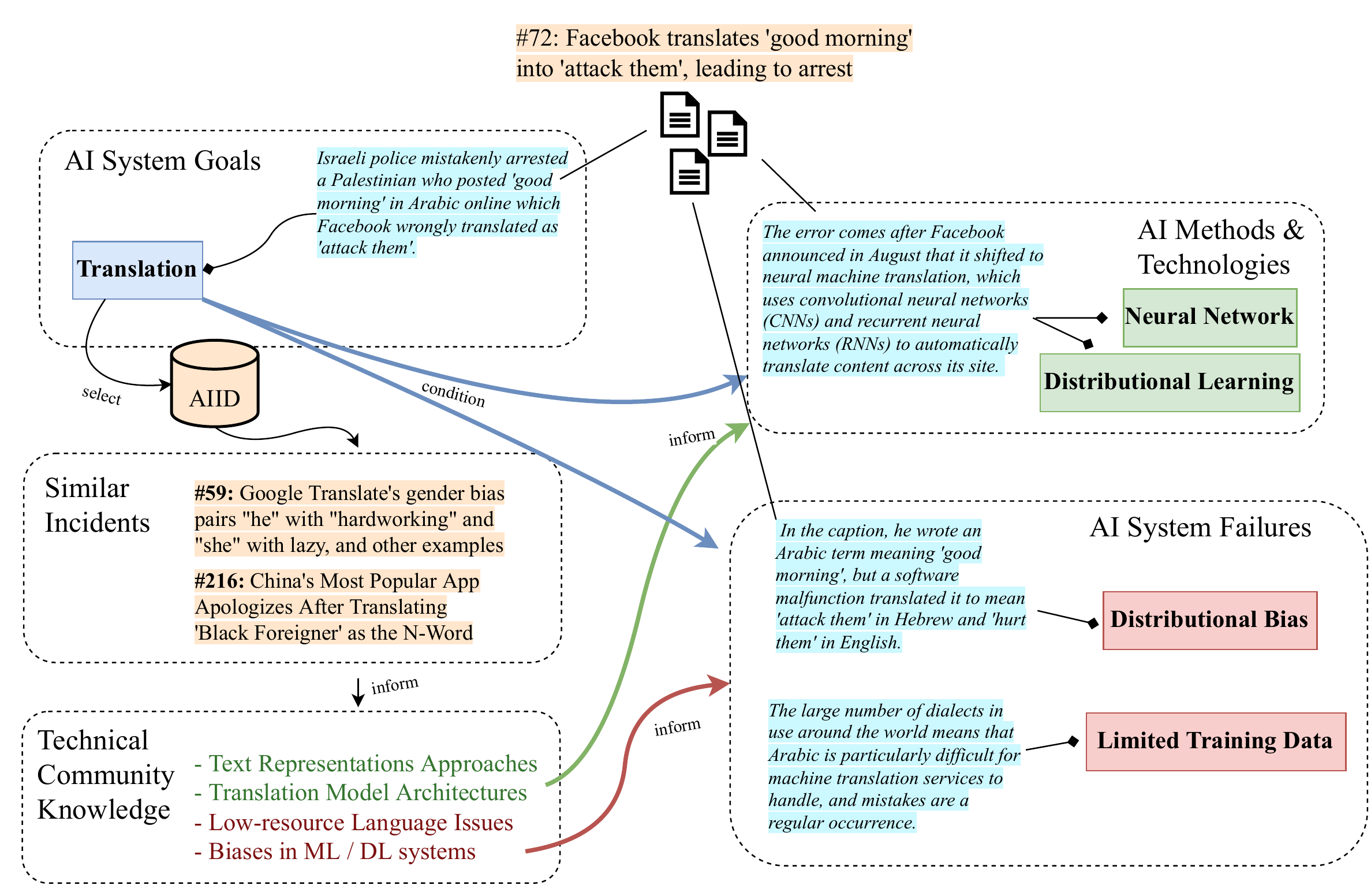} 
 \caption{An example of the proposed GMF classification workflow for the real-world AIID incident \#72. Incidents titles and report passages are illustrated with an orange and cyan highlight respectively, while colored rectangles represent GMF classifications and diamond edges showcase their grounding. The ``Translation'' AI System Goal is directly attainable from relevant passages, which enables similar incident retrieval from AIID; these are utilized to obtain technical knowledge about translation methods, models, implementations as well as known limitations and issues, from experts in the AI / ML / Safety community. This feedback in turn informs and introduces priors to AI Methods and Technologies and AI Failure Causes, in conjunction with useful incident contents. Comments and some interactions are omitted from the visualization for readability.}
  \label{fig:example}
\end{figure*}


\subsection{Development}\label{sec:development} 

We now present the proposed GMF development process, which adopts an iterative, bottom-up approach, working from batches of individual incidents from AIID \cite{mcgregor2021preventing} to annotate incidents and populate taxonomy contents.
In this setting, the proposed taxonomic system solves an epistemological problem for AIID through an open tagging design that interrelates both high-certainty and speculative classifications. Tags are open in the sense that there is no pre-defined set of goals, methods, or failures so these can develop through time and reflect consensus in the AI Safety research community.

Given an AI incident, the general proposed workflow for an annotator is as follows:
\begin{enumerate}
\item Read through incident contents and identify salient passages, e.g. text mentioning technical terms, system use cases, specifications and harms.
\item Pair salient passages with free discussion comments, providing explanation, rationale, additional information and linkage to external resources, if deemed necessary.
\item Create / modify a GMF taxonomy classification $C$, following the taxonomy definitions, workflow protocols and information sources established in Sections \ref{sec:taxonomies} and \ref{sec:workflow}. Afterwards, link one or more salient passages to the classification, such that they provide reasonable justification and grounding to selecting $C$.
\item Pair each classification with a confidence modifier, i.e. ``known'' or ``potential'', conveying near-certain or above-average degrees of certainty that $C$ is relevant to the incident. The totality of accumulated information (selected snippet, content terms / technical information / ambiguity, historical incidents, technical background and community knowledge) should determine the most relevant modifier according to the annotator.
\item Pair $C$ with free discussion comments, which supply adequate reasoning for why the classification and confidence modifier are fitting / relevant to the linked passages, given the totality of accumulated information available to the annotator. This discussion should be able to reveal the decision-making process, rationale and evidence used, serving as documentation to other interested third parties (annotators, evaluators, editors etc.). Notably, such comments are especially important in annotations where non-trivial amounts of experience, intellectual work and information gathering were marshalled to produce $C$, which are classification cases that will be characterized by higher uncertainty, on average.
\end{enumerate}

An illustration of the overall GMF structure, workflow and development protocol for classification of a real-world incident \footnote{\url{https://incidentdatabase.ai/cite/72}} is available in Figure \ref{fig:example}.

The proposed annotation configuration provides a number of notable desired features to the taxonomy development process.
First, improved transparency and validation of classifications is achieved by grounding annotations with relevant passages and free discussion comments. Passages list supporting input evidence, while comments may elaborate on rationale, sources, reasoning and intuition across different annotators, levels of expertise and points in time.

Second, the linkage stated above comes with built-in potential for data-driven automation, enabling the development of Machine Learning workflows to enhance, automate and accelerate future manual annotation efforts via, e.g., classification recommendations, salient passage extraction, keyword extraction, etc.

Finally, the proposed workflow can rapidly generate annotated dataset versions of variable levels of classification grounding and quality control; given the large cost of manual technical annotation, quality can be improved by iteratively refining existing versions of ground truth. For example, an initial release might include noisy crowd-sourced annotations (e.g. with limited classification grounding and a small annotator pool). Subsequent versions that undergo multiple steps of correction, verification and validation should produce datasets eligible for research-grade utilization (e.g. via imposing minimum levels / thresholds for classification grounding statistics, annotator pool size and agreement, etc.).

We have currently conducted
a series of classification exercises on a representative set of incidents, to explore the information and context available to back various taxonomy designs;
a summary of the status of GMF ontologies at this stage is provided in table \ref{tab:status}, while
findings from this investigation are consolidated in the interrelated structure and workflows herein introduced.
We have iterated the taxonomy system over the first $\approx 10 \%$ percent of the database
with expert annotators (i.e. ML / AI Safety researchers and engineers
at a PhD level) and are now prepared to comprehensively apply the
taxonomy across the database on an ongoing basis.

At the same time, we are working on developing, designing and integrating dedicated toolsets, interfaces and procedures to support crowdsourced operations in the near future; namely, the proposed developmental workflow will be supported by a dedicated user interface designed to facilitate fast-paced annotation and support automation functionalities (e.g. auto-completion, recommendation, highlighting, etc.) for improved user experience and reduced boilerplate.
Apart from annotation, this infrastructure will support information extraction, incident retrieval and navigation for the inspection of possible failure modes, causes and risks in existing AI incidents, as well as exploratory analysis on deployment descriptions of new systems, by experts and laypeople alike.

\begin{table}
  \centering
  \begin{tabular}{lr}
    Statistic & count \\ \toprule
    Number of annotated AIID incidents & 41 \\
    AIID coverage \% & 12.40 \\ \midrule
    \textit{Known} annotations \% & 48.68  \\
    \textit{Potential} annotations \% & 51.32  \\  \midrule
    Goals per incident & 1.20 \\
    Methods / Technologies per incident & 2.43 \\
    Technical Failure Causes per incident & 3.78 \\
    \bottomrule
  \end{tabular}
  \caption{Status of initial GMF development, with respect to AI incident instances in AIID.}
  \label{tab:status}
\end{table}

\subsection{Expected Impact}\label{sec:impact}
The proposed taxonomic system will complement existing ontologies by providing applicability to a broad set of real-world AI incidents (and textual content of AI misbehavior in general), with a strong a focus on rigorous, technical descriptions of failure causes.
We expect the set of annotation workflows, tools and resources will enable rapid GMF annotation, curation and verification, leverage the support of AI communities and experts alike, and generate a wide variety of useful data-driven applications for further automation and development of related research.

At the current early stage of GMF construction, we center our offering on real-world AI harm events and posit that the proposed body of work can help researchers identify, analyse and contextualize open problems in AI Safety and related research domains (e.g. Human-Centered AI, Cybersecurity, Machine Ethics, etc), aid policymakers in effectively regulating the most damaging systems, and empower corporations identify when systems under development are subject to previously experienced failure modes.

\section{Conclusion and Future Work}

In this work, we presented preliminary work on the GMF taxonomic system, a proposed set of taxonomies that capture AI System Goals, AI Methods and Technologies and AI Failure Causes. We presented recommendations for a taxonomy annotation workflow, development procedure and future plans, in the context of incident reports in AIID, listing the rationale, benefits and expected impact of the resulting resources to the research, policy and industry sectors.
At present, we have applied classifications from the perspective of machine learning research engineers, under an iterative taxonomy development procedure on an initial batch of incident data.
Given that incidents are typically multi-faceted in their causes and safe system design calls for a variety of organizational processes in addition to design accommodations,
in the future we plan to augment the classification set with additional perspectives (e.g., scientists and researchers with expertise in philosophy, ethics, AI governance and various human factors). Additionally, we plan on publishing the taxonomy within the platform interface of AIID \footnote{\url{https://incidentdatabase.ai}} and invite external participation in the incident classification and taxonomy development process in a crowdsourced setting.
We believe that the broad inclusion of AI Safety-related research communities will provide invaluable feedback, constructive revisions and significant expansions to existing annotations and ontology contents, lending critical technical insights about the holistic description of AI system failures, towards building up mitigation measures of future AI-induced harms in the real world.

\textbf{Acknowledgements}

In addition to participating in the 2018 discussions that launched the AIID, Richard Mallah gave valuable feedback on the taxonomy. The AIID is an effort by many people and organizations operating under the banner of the Responsible AI Collaborative. It is through their collective efforts that the ontological perspectives presented above have meaning and real world importance.

\bibliographystyle{acm}
\bibliography{references}

\end{document}